# OpenVIO: A Versatile Open-Source Visual-Inertial Odometry System

Seungwon Choi

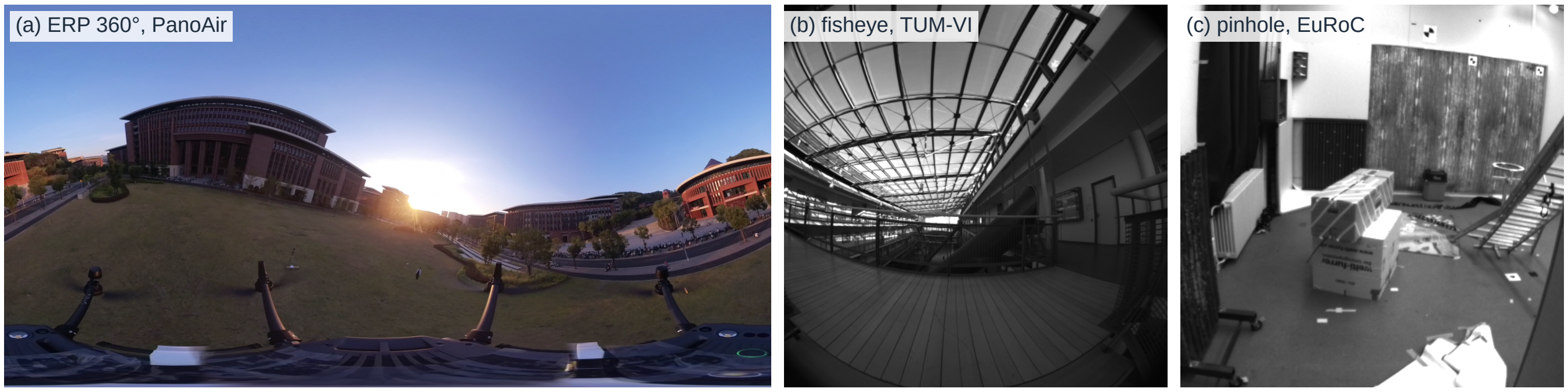

Fig. 1. Example images for the camera models supported by OpenVIO. (a) Stitched 360° equirectangular (ERP) image from PanoAir [1]. (b) Fisheye image from TUM-VI [2]. (c) Pinhole image from EuRoC [3].

***Abstract*— Visual-inertial odometry (VIO) is a fundamental topic in robotics research. Beyond research, VIO has been adopted in a wide range of products. In this manuscript, we present OpenVIO, an open-source VIO system for research and practical applications. It is released under the MIT license. OpenVIO supports pinhole and fisheye cameras and extends this support to stitched 360° images. We describe each part of the system together with its design considerations. We also report benchmark results with other open-source VIO systems on public datasets. OpenVIO is available as open source at `https://github.com/93won/openvio`.**

## I. Introduction

Visual-inertial odometry (VIO) combines camera images with measurements from an inertial measurement unit (IMU) to estimate ego-motion. This cost-effective sensor combination supports reliable motion estimation and has made VIO useful in products such as mobile robots, drones, augmented reality (AR) devices and autonomous cars [4]–[7].

In this manuscript, we present OpenVIO, an open-source VIO system with stable performance across diverse camera models. It supports pinhole and fisheye cameras, as well as stitched 360° images with a full spherical field of view (Fig. 1). Under the MIT license, it can be used in both research and products. We compare OpenVIO with nine open-source VIO systems on EuRoC and TUM-VI in monocular and stereo configurations. We also evaluate OpenVIO on the 17 flights of PanoAir.

The rest of this manuscript is organized as follows. Section II reviews related work. Section III describes the OpenVIO pipeline, including the front end, initialization, filtering and extensions for 360° cameras. Section IV reports the benchmark results on EuRoC, TUM-VI and PanoAir. Section V concludes the manuscript.

TABLE I

Open-source VIO systems: licenses and supported cameras

| System | License | Camera models | Configuration |
|---|---|---|---|
| MSCKF-VIO [8] | Penn research | pinhole, fisheye | stereo |
| ROVIO [9] | BSD-3 | pinhole, fisheye | mono |
| OKVIS [10] | BSD-3 | pinhole, fisheye | mono, stereo |
| VINS-Mono [11] | GPL-3.0 | pinhole, fisheye | mono |
| VINS-Fusion [12] | GPL-3.0 | pinhole, fisheye | mono, stereo |
| SVO Pro [13] | GPL-3.0 | pinhole, fisheye | mono, stereo |
| OpenVINS [14] | GPL-3.0 | pinhole, fisheye | mono, stereo |
| Basalt [15] | BSD-3 | pinhole, fisheye | stereo |
| Kimera-VIO [16] | BSD-2 | pinhole, fisheye | mono, stereo |
| **OpenVIO** | **MIT** | pinhole, fisheye, **360° ERP** | mono, stereo |

## II. Related Work

Over the past decade, many open-source VIO systems have been released (Table I). Based on their state estimation methods, these systems fall into two main groups: filter-based and optimization-based approaches. Filter-based systems, such as MSCKF-VIO [8], ROVIO [9] and OpenVINS [14], combine IMU propagation with visual measurement updates in an extended Kalman filter. Optimization-based systems, such as OKVIS [10], VINS-Mono [11], VINS-Fusion [12], SVO Pro [13], Basalt [15] and Kimera-VIO [16], solve a nonlinear least-squares problem over a sliding window of keyframes. These open-source systems have supported further research and the development of practical applications.

OpenVIO uses a filter-based approach. It tracks features and updates a sliding window of past IMU poses with an extended Kalman filter.

## III. OpenVIO

This section describes the OpenVIO pipeline, its main components, and the extensions for stitched 360° cameras.

S. Choi, `csw3575@snu.ac.kr`. The code is available under the MIT license at `https://github.com/93won/openvio`.

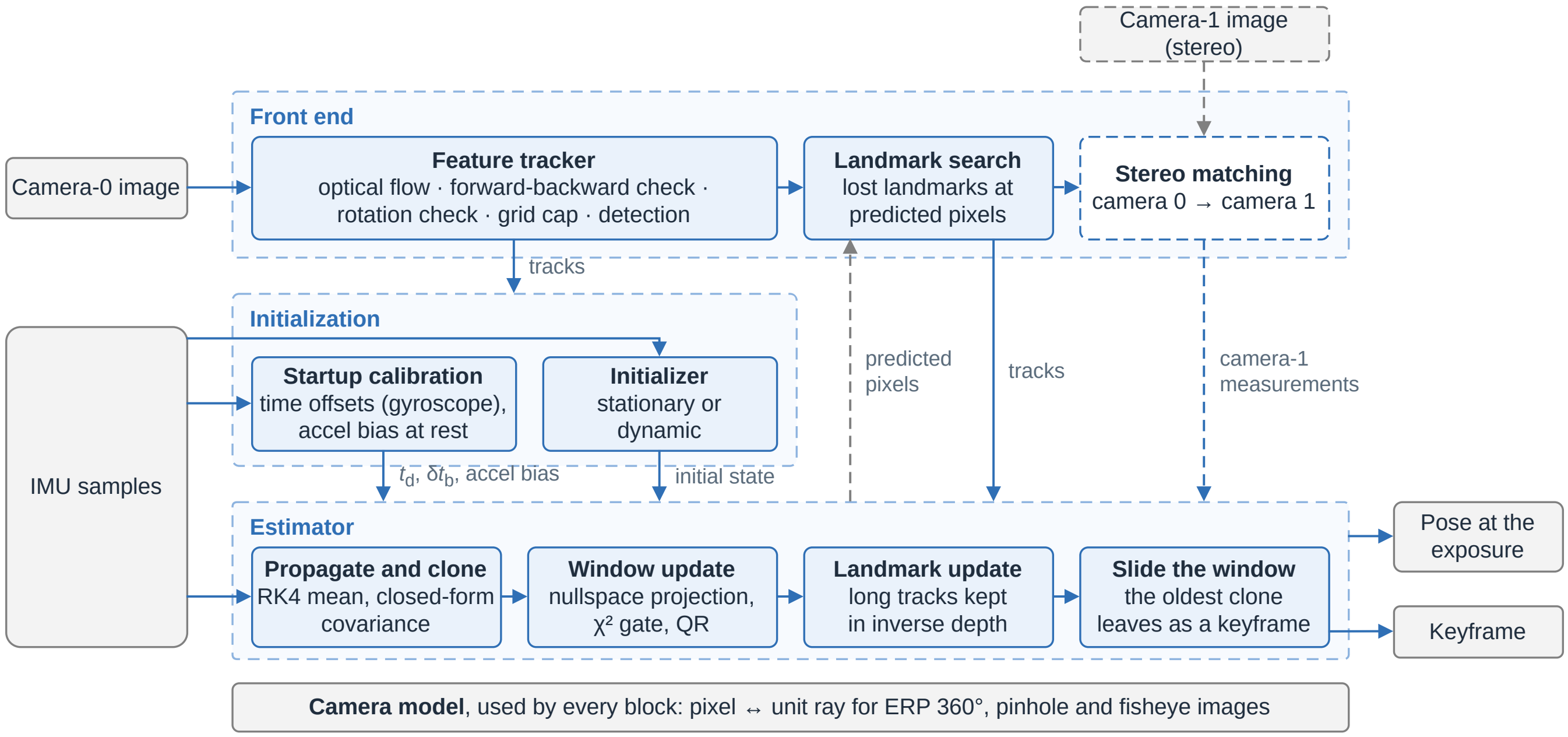

Fig. 2. Processing of one image in OpenVIO. The dashed camera-1 and stereo-matching boxes indicate stereo-only steps. The gray dashed feedback arrow carries predicted pixels to landmark search. Pale dashed borders group processing stages.

### A. Overview

OpenVIO processes monocular or stereo images and IMU samples (Fig. 2). The front end tracks features for initialization and subsequent filter updates. OpenVIO outputs the IMU pose at each image exposure. When a pose leaves the filter window, OpenVIO exports it as a keyframe with its feature observations.

Each camera model converts a pixel into a unit ray, a unit-length direction vector in the camera frame. The filter uses rays for geometric operations and the camera model's projection for image residuals. This interface supports directions behind the camera and lets the same estimator handle all camera models in Table I.

### B. Front End

The front end tracks features by pyramidal Lucas–Kanade optical flow [17], [18] with a forward-backward check, which rejects a match when tracking it back gives a pixel too far from its original position. When auto exposure changes the brightness, the previous image is scaled by the ratio of the two median intensities before optical flow.

A RANSAC [19] rotation fit between unit rays [20] provides a further outlier check. The fit checks angular agreement across tracks for every camera model. The filter subsequently estimates both rotation and translation. Each grid cell keeps a limited number of features, with priority given to longer tracks. This preserves observation histories while preventing features from concentrating in a small image region. New corners [21] fill empty cells.

A landmark that the tracker loses can be found again. Each landmark stores a patch from its most recent accepted observation. The front end matches this patch by normalized cross-correlation around the pixel predicted by the filter. The patch is sampled on the tangent plane of the landmark's ray to account for lens distortion and ERP image stretching. For ERP images, this sampling continues across the left and right image edges. A recovered match retains the landmark's identity and must still pass the filter's measurement check.

For stereo, optical flow also matches each feature into the second image. The initial pixel is predicted by rotating and projecting its ray into the second camera. This assumes a point at infinity, so the prediction uses only rotation. Optical flow then finds the actual correspondence, including its disparity. The second observation enters the filter with the calibrated translation between the cameras.

### C. Initialization

Initialization supplies the IMU orientation, velocity, biases and covariance needed by the filter. OpenVIO buffers feature tracks and IMU samples, and selects stationary or dynamic initialization based on motion estimated by feature tracking. The initializer uses tracks from the first camera (camera 0) in both monocular and stereo modes. It compares the mean feature displacement in the older and newer halves of the buffer. A stationary interval followed by motion triggers stationary initialization; a moving interval uses dynamic initialization. If both halves remain stationary, it waits for motion.

*Stationary initialization.* The accelerometer variation must also confirm the transition from rest to motion. The mean gyroscope reading of the still interval gives the gyroscope bias. The mean accelerometer reading defines the initial gravity direction, and its magnitude relative to the known gravity magnitude gives the accelerometer bias component along that direction. Position and velocity are initialized to zero at the end of the still interval. Buffered IMU samples then propagate this state through the subsequent motion.

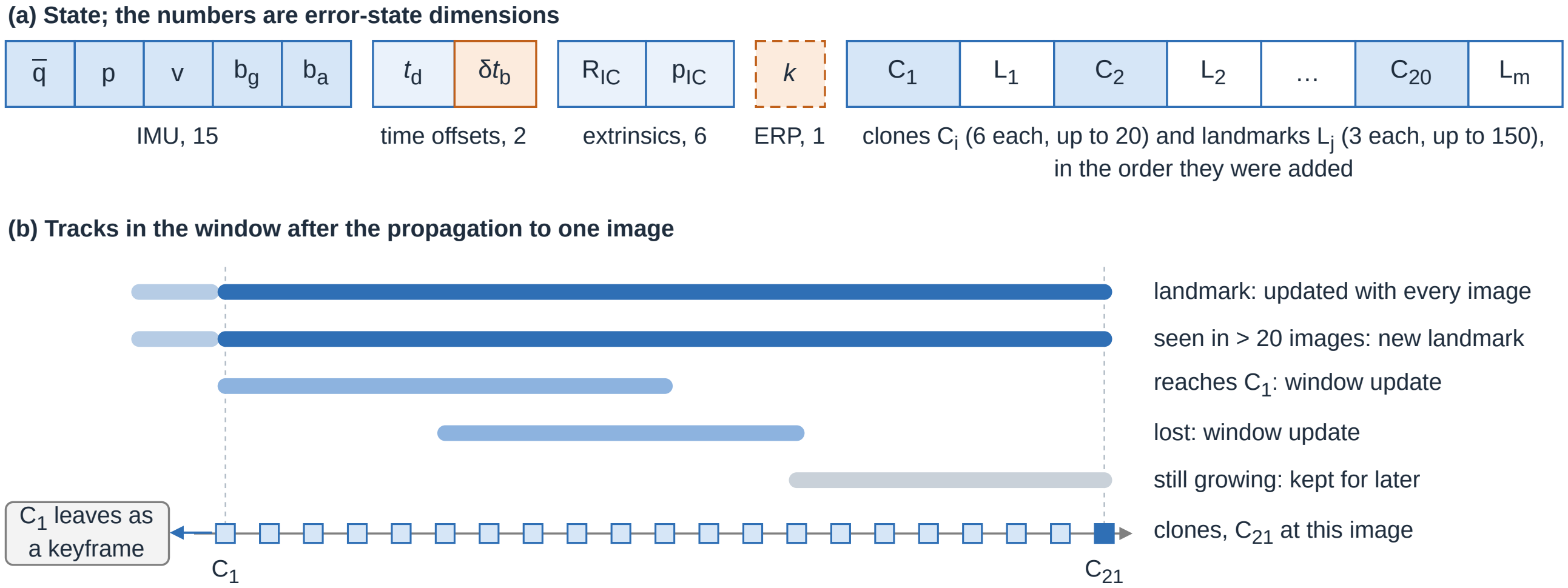

Fig. 3. State and window. (a) Clones and landmarks appear in their order of addition. Orange marks two-lens 360° camera states; the dashed box marks the online ERP parameter. (b) Lost tracks and tracks reaching the oldest clone update the window. Accepted long tracks become landmarks, updated at each observation. The oldest clone leaves as a keyframe.

*Dynamic initialization.* OpenVIO first solves the linear initialization problem described in [22]. The unknown vector $\mathbf{z}$ contains the feature positions $\{\mathbf{p}_{F_j}\}$, initial velocity $\mathbf{v}_0$ and gravity vector $\mathbf{g}_0$, expressed in the first IMU frame. Unit-ray observations and IMU preintegration form $\mathbf{A}$ and $\mathbf{b}$, giving

$$\hat{\mathbf{z}} = \arg\min_{\mathbf{z}} \|\mathbf{A}\mathbf{z} - \mathbf{b}\|^2, \qquad \text{subject to } \|\mathbf{g}_0\| = g, \tag{1}$$

where $g$ is the known gravity magnitude. The camera–IMU extrinsics and bias priors are fixed in this linear step. The initializer requires enough tracked features, IMU samples and accumulated rotation.

The linear solution seeds a joint refinement of the poses, velocities, IMU biases and feature positions, collected in $\mathcal{X}$. Ceres [23] minimizes

$$\hat{\mathcal{X}} = \arg\min_{\mathcal{X}} \Bigg[ \|\mathbf{r}_{\text{prior}}\|^2 + \sum_k \|\mathbf{r}_{\text{IMU},k}\|^2 + \sum_{k,j} \rho\left(\|\mathbf{r}_{\text{image},kj}\|^2\right) \Bigg]. \tag{2}$$

The residuals are normalized by their uncertainty. They comprise a prior on the first pose and biases, IMU constraints between consecutive poses, and reprojection errors of feature $j$ at pose $k$. The prior sets the position and yaw reference, while the Cauchy loss $\rho$ limits the influence of image outliers. After convergence, the newest IMU state and its marginal covariance initialize the filter. Selected covariance blocks are inflated to avoid an overconfident start. If refinement or covariance recovery fails, OpenVIO collects more data and tries again.

*Startup calibration.* Before normal filter updates, OpenVIO uses the buffered data to seed the enabled time-offset states. Camera rotations from feature tracks are compared with gyroscope rotations over time-shifted intervals, with a constant gyroscope bias accounted for. For a two-lens ERP camera, the difference between the two estimated offsets gives the back-lens exposure offset (Section III-E). A time estimate is accepted only when the data constrain it sufficiently; otherwise, the correction starts at zero and can be estimated online.

When the recording starts at rest, an optional correction further refines the accelerometer bias component along gravity. This correction combines the mean accelerometer readings over the buffered interval with the initialized vertical velocity.

### D. Filter

OpenVIO combines a multi-state constraint Kalman filter (MSCKF) [24] with persistent landmarks in its state [25]. This uses constraints from completed tracks and repeated observations of selected landmarks. Grouped by variable type, the state is

$$\mathbf{x} = \left[\, \mathbf{x}_I^\top,\ t_d,\ \delta t_b,\ \mathbf{x}_{\text{cal}}^\top,\ \mathbf{x}_{C_1}^\top, \ldots, \mathbf{x}_{C_n}^\top,\ \mathbf{x}_{L_1}^\top, \ldots, \mathbf{x}_{L_m}^\top \,\right]^\top, \tag{3}$$

with the IMU state $\mathbf{x}_I$ (orientation, position, velocity and the gyroscope and accelerometer biases), the time offset $t_d$, the exposure delay $\delta t_b$ of the back lens of a 360° camera, the calibration $\mathbf{x}_{\text{cal}}$ (camera–IMU extrinsics and the stitching error $k$ of the ERP model), the clones $\mathbf{x}_{C_i}$ of past IMU poses and the landmarks $\mathbf{x}_{L_j}$ in inverse depth [26]. Figure 3a shows the order of addition; Eq. (3) groups variables by type.

The filter propagates the IMU state with fourth-order Runge–Kutta integration and the covariance with the closed-form method of [27], using first-estimate Jacobians [28]. At each image stamped $t$, the IMU pose at its exposure $t + t_d$ is added as a clone [29]. The clone preserves the pose and its correlations with the rest of the state, so later observations can constrain the motion between image exposures. Image residuals use current estimates, while the associated Jacobians use the retained first estimates.

Tracks are triangulated from unit rays, with the point closest to all rays as the initial estimate. For inverse-depth refinement [24], the anchor camera frame and each observation's camera frame are rotated so that their respective measured rays align with the local $+z$ axis. This parameterization handles rays in any direction, including those behind an ERP camera.

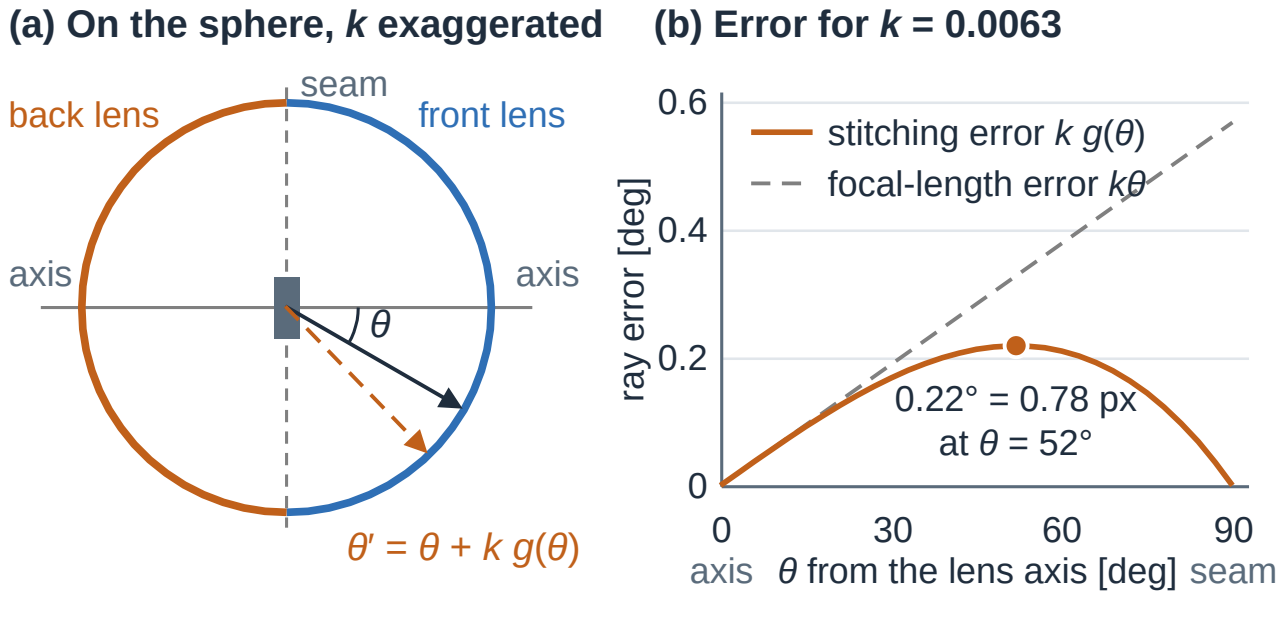

Fig. 4. Stitching error of a two-lens ERP image. (a) A ray at the angle $\theta$ from its lens axis is placed at $\theta' = \theta + k\,g(\theta)$ ($k$ exaggerated). (b) The error for $k = 0.0063$, the prior on PanoAir; the pixel equivalent is at the ERP equator.

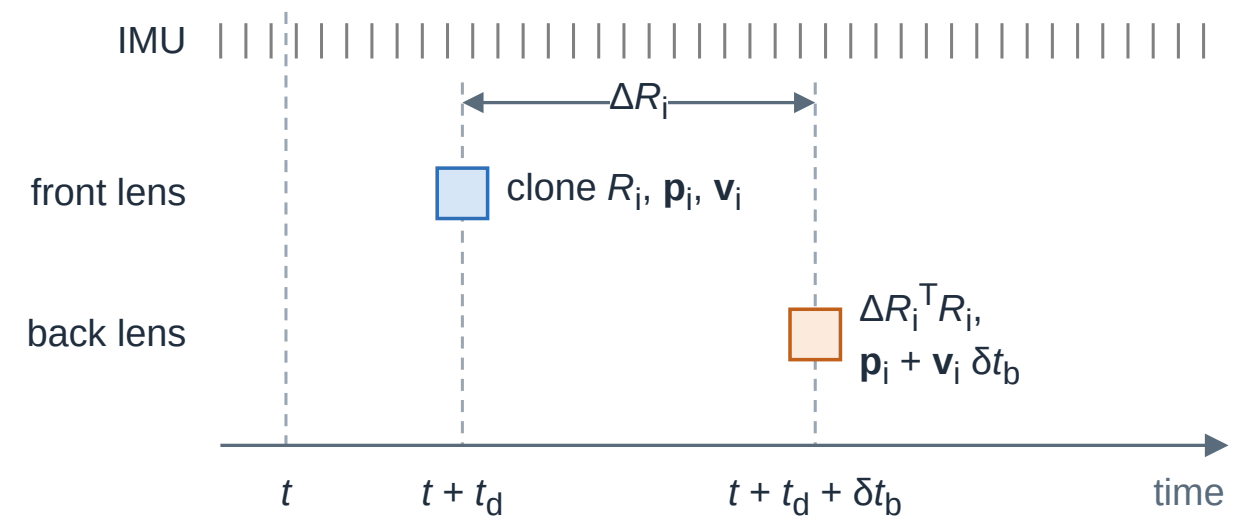

Fig. 5. Timing of one stitched image. The front and back lenses are exposed at $t + t_d$ and $t + t_d + \delta t_b$, respectively. The diagram shows a positive $\delta t_b$; back-lens rays use the pose at their own exposure.

A track that is lost or reaches back to the oldest clone updates the window (Fig. 3b). Its point is triangulated, and nullspace projection removes the point's contribution from the linearized residuals [24]. Tracks that pass a $\chi^2$ gate are combined into one update and compressed by QR decomposition. The resulting constraint updates the state without retaining the triangulated point as a state variable. Once consumed, those track observations are removed to prevent their reuse.

A track seen in more images than the window holds is a candidate for a persistent landmark. It must pass triangulation and the measurement gate, and space must remain in the landmark budget. An optional depth-uncertainty test also prevents poorly constrained points from entering the state; such tracks can still contribute a window update. Each accepted landmark retains its inverse depth and correlations with the other states, allowing new observations to constrain motion after the original clones leave the window. Shorter tracks that are still growing wait for more observations.

After the update, the oldest clone leaves as a keyframe when the window exceeds its limit. The released configurations keep up to 20 clones and 150 landmarks. The time offset and camera–IMU extrinsics are estimated online.

A landmark whose track is lost stays in the state for up to 5 s, so that the front end can find it again. It is removed when its inverse depth is not positive, when the gate rejects it twice in a row, or when its inverse depth departs too far from its first estimate, to limit scale errors from inaccurate first-estimate Jacobians.

### E. 360° Cameras

OpenVIO takes an ERP image already stitched by a two-lens 360° camera. The two back-to-back fisheye lenses cover the sphere, and the ERP image maps the longitude of a ray to its column and the latitude to its row:

$$u = W\left(\frac{1}{2} + \frac{\operatorname{atan2}(x,z)}{2\pi}\right), \qquad v = H\left(\frac{1}{2} + \frac{\arcsin(y/\|\mathbf{p}\|)}{\pi}\right). \tag{4}$$

Here, $W$ and $H$ are the image width and height in pixels. The ray $\mathbf{p} = (x, y, z)^\top$ uses camera axes pointing right ($+x$), down ($+y$) and forward ($+z$). The front lens sees the rays with $z > 0$ and the back lens the others. For such a camera OpenVIO estimates two more states, the stitching error $k$ and the exposure delay $\delta t_b$ of the back lens (Fig. 3a). The first corrects the pixel-to-ray geometry, while the second determines which pose predicts a back-lens observation. Both are estimated within the same filter as the trajectory.

*Stitching error.* The stitching relies on the camera's own calibration, and a small error in it places rays at slightly wrong angles from their lens axis. For a ray at angle $\theta$ from that axis, OpenVIO models the error with one parameter $k$ (Fig. 4),

$$\theta' = \theta + k\,g(\theta), \qquad g(\theta) = \theta\left(1 - \left(\tfrac{2\theta}{\pi}\right)^2\right). \tag{5}$$

The same angular correction is applied about each lens axis without changing the ray's azimuth. The correction $k\,g(\theta)$ vanishes at the lens axis and at the seam, so these directions remain fixed. Near the axis, it grows like $k\theta$, as a focal-length error would. Thus, one state captures a smooth angular mismatch in the stitched image. For $k = 0.0063$, the prior on PanoAir, the correction peaks at 0.22° at $\theta = 52°$, equivalent to 0.78 px along the ERP equator at an image width of 1280 pixels (Fig. 4b). The filter estimates $k$ online.

*Exposure delay.* A stitched image can combine two lens exposures at different times (Fig. 5). During motion, each half then corresponds to a different pose. The common camera–IMU offset $t_d$ sets the front exposure $t + t_d$, and the relative offset $\delta t_b$ sets the back exposure $t + t_d + \delta t_b$. OpenVIO retains the front-exposure pose as a clone and predicts the back-exposure pose with constant linear velocity between exposures,

$$R_{i,b} = \Delta R_i^\top R_i, \qquad \mathbf{p}_{i,b} = \mathbf{p}_i + \mathbf{v}_i\,\delta t_b, \tag{6}$$

Here, $R_i$ maps global vectors into the IMU frame at the front exposure, while $\mathbf{p}_i$ and $\mathbf{v}_i$ are expressed in the global frame. The gyroscope-derived $\Delta R_i$ maps vectors from the back-exposure IMU frame into the front-exposure IMU frame. The filter refines $\delta t_b$ online, allowing either sign.

## IV. Benchmark

### A. Setup

We compare OpenVIO with the nine open-source VIO systems in Table I on EuRoC MAV [3] and TUM-VI [2],

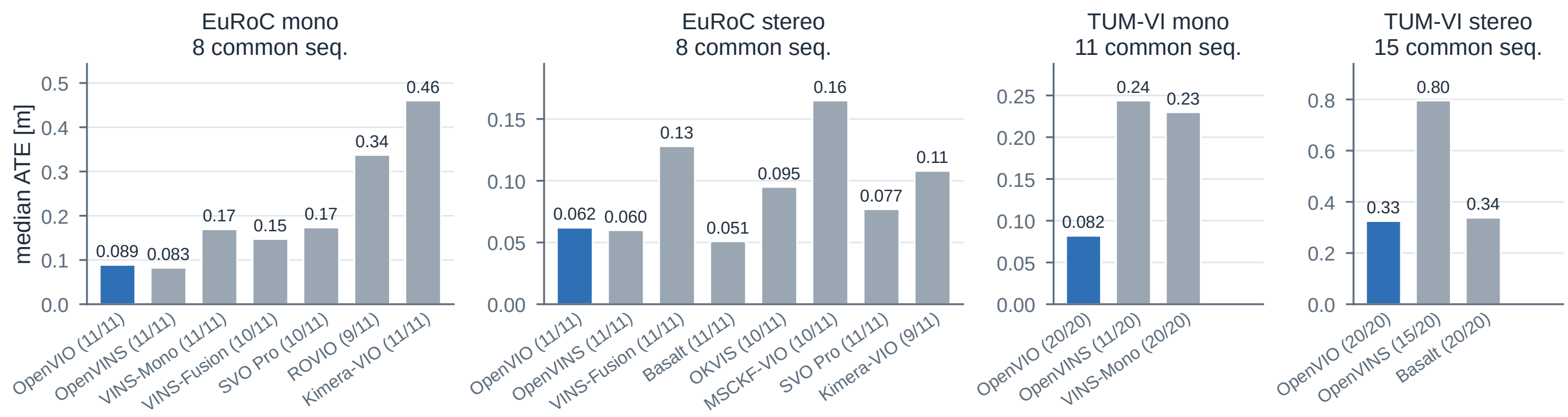

Fig. 6. Median of the per-sequence ATE values over the sequences that all systems in each panel complete. The underlying values are the medians across runs in Tables II, III and IV. Parentheses show completed sequences out of all evaluated sequences.

TABLE II

EuRoC, monocular. ATE RMSE [m] after SE(3) alignment, median of five runs. ×: at least half the runs failed. The mean uses the eight sequences completed by all systems. Best per row in bold.

| Sequence | OpenVIO | OpenVINS | VINS-Mono | VINS-Fusion | SVO Pro | ROVIO | Kimera-VIO |
|---|---|---|---|---|---|---|---|
| MH_01 | 0.168 | **0.096** | 0.192 | 0.186 | 0.438 | 0.455 | 1.352 |
| MH_02 | **0.076** | 0.152 | 0.167 | 0.088 | 0.094 | × | 0.280 |
| MH_03 | **0.118** | 0.124 | 0.217 | 0.135 | 8.488 | 0.486 | 0.471 |
| MH_04 | **0.110** | 0.223 | 0.378 | 0.217 | × | 1.003 | 5.282 |
| MH_05 | 0.429 | 0.624 | 0.354 | 0.326 | **0.216** | 1.164 | 6.148 |
| V1_01 | 0.061 | 0.069 | 0.079 | 0.059 | **0.053** | 0.118 | 2.555 |
| V1_02 | 0.064 | **0.058** | 0.098 | 0.086 | 0.060 | 0.184 | 0.100 |
| V1_03 | **0.060** | 0.066 | 0.147 | 0.160 | 0.131 | 0.201 | 0.213 |
| V2_01 | **0.076** | 0.104 | 0.085 | × | 0.080 | × | 0.290 |
| V2_02 | **0.067** | **0.067** | 0.122 | 0.119 | 0.131 | 0.743 | 0.198 |
| V2_03 | **0.111** | 0.178 | 0.289 | 0.208 | 0.248 | 0.220 | 0.450 |
| Mean (8 seq.) | 0.135 | 0.160 | 0.187 | 0.160 | 1.221 | 0.446 | 1.436 |
| Completed | 11/11 | 11/11 | 11/11 | 10/11 | 10/11 | 9/11 | 11/11 |

using the available monocular and stereo configurations. EuRoC contains 11 drone sequences recorded with a stereo pinhole camera. We use the 20 indoor TUM-VI sequences, recorded with a hand-held stereo fisheye camera at 512×512 pixels. Each baseline uses its authors' configuration with loop closure disabled. We omit a system from a dataset when its authors provide no configuration for it. OpenVIO uses its released configurations, with the parameter variations described below.

We report absolute trajectory error (ATE) as the root-mean-square error (RMSE) of positions after rigid SE(3) alignment [30]. For TUM-VI, only poses with motion-capture ground truth enter the alignment and error calculation. On sequences that leave and return to the room, this measures accumulated drift at the observed portions rather than error along the unobserved route.

Each table entry is the median of five runs for EuRoC and TUM-VI room sequences, or three runs for the remaining TUM-VI sequences. A run fails if no valid ATE is available, its ATE exceeds 10 m, or its evaluated time span covers less than half the sequence duration. Failed runs enter the median as infinite error. A sequence is counted as completed when more than half its runs succeed.

OpenVIO is deterministic, so its repeated runs use small parameter variations. The first three runs use the released configuration and two variants that scale both gyroscope and accelerometer noise densities by 0.9 or 1.1. For the five-run sets, two further runs retain the default IMU parameters and change the optical-flow termination tolerance from 0.01 to 0.0099 and 0.0098. The recorded sensor measurements remain the same in every run.

### *B. Accuracy on EuRoC and TUM-VI*

Tables II, III and IV report the per-sequence results. For each dataset and camera configuration, aggregate errors use only sequences completed by all compared systems. Figure 6 shows the median across these sequences, while the table summaries show the mean. Completion counts cover all evaluated sequences.

OpenVIO completes all EuRoC and TUM-VI sequences in both monocular and stereo modes under the criterion above. On EuRoC, its median ATE is close to OpenVINS monocular and to Basalt and OpenVINS stereo. On TUM-VI, OpenVIO has the lowest median ATE over the common sequences in both configurations. These results show consistent sequence completion with competitive accuracy across the two datasets.

TABLE III

EuRoC, stereo. ATE RMSE [m] after SE(3) alignment, median of five runs. ×: at least half the runs failed. The mean uses the eight sequences completed by all systems. Best per row in bold.

| Sequence | OpenVIO | OpenVINS | VINS-Fusion | Basalt | OKVIS | MSCKF-VIO | SVO Pro | Kimera-VIO |
|---|---|---|---|---|---|---|---|---|
| MH_01 | 0.091 | 0.070 | 0.253 | **0.066** | 0.213 | × | 0.120 | 0.199 |
| MH_02 | 0.069 | 0.111 | 0.217 | **0.055** | 0.120 | 0.178 | 0.065 | 0.197 |
| MH_03 | 0.075 | 0.144 | 0.326 | **0.062** | 0.135 | 0.226 | 0.077 | 0.321 |
| MH_04 | **0.091** | 0.155 | 0.434 | 0.114 | 0.180 | 0.223 | 0.176 | × |
| MH_05 | 0.210 | 0.441 | 0.307 | **0.145** | 0.297 | 0.345 | 0.182 | 0.257 |
| V1_01 | 0.049 | 0.062 | 0.113 | **0.043** | 0.046 | 0.086 | 0.056 | 0.058 |
| V1_02 | **0.042** | 0.048 | 0.108 | 0.045 | 0.054 | 0.112 | 0.077 | 0.058 |
| V1_03 | **0.049** | 0.058 | 0.112 | 0.053 | 0.110 | 0.188 | 0.075 | 0.090 |
| V2_01 | 0.068 | 0.050 | 0.132 | **0.039** | 0.056 | 0.084 | 0.091 | 0.059 |
| V2_02 | 0.056 | **0.035** | 0.124 | 0.049 | 0.080 | 0.152 | 0.135 | 0.126 |
| V2_03 | 0.125 | **0.085** | 0.318 | 0.229 | × | 1.940 | 0.202 | × |
| Mean (8 seq.) | 0.077 | 0.119 | 0.180 | 0.061 | 0.112 | 0.171 | 0.095 | 0.146 |
| Completed | 11/11 | 11/11 | 11/11 | 11/11 | 10/11 | 10/11 | 11/11 | 9/11 |

TABLE IV

TUM-VI 512, monocular and stereo. ATE RMSE [m] after SE(3) alignment, median of the runs (5 per room, 3 otherwise); ×: half or more of the runs failed. Best per row and configuration in bold. The means are over the sequences that all systems of a configuration complete (monocular 11, stereo 15).

| | Monocular | | | Stereo | | |
|---|---|---|---|---|---|---|
| Sequence | OpenVIO | OpenVINS | VINS-Mono | OpenVIO | OpenVINS | Basalt |
| room1 | **0.033** | 0.048 | 0.056 | **0.034** | 0.056 | 0.106 |
| room2 | **0.044** | 0.082 | 0.056 | **0.041** | 0.088 | 0.076 |
| room3 | 0.079 | **0.075** | 0.149 | **0.064** | 0.080 | 0.133 |
| room4 | 0.034 | **0.027** | 0.044 | 0.028 | **0.027** | 0.052 |
| room5 | **0.065** | 0.075 | 0.216 | **0.068** | 0.096 | 0.180 |
| room6 | **0.052** | × | 0.077 | 0.049 | × | **0.023** |
| corridor1 | **0.679** | 0.817 | 0.987 | 0.241 | 0.796 | **0.207** |
| corridor2 | **0.212** | 0.244 | 1.248 | **0.341** | 0.403 | 0.412 |
| corridor3 | **0.346** | 1.092 | 1.435 | **0.381** | 1.053 | 0.493 |
| corridor4 | **0.082** | 0.391 | 0.230 | **0.121** | 0.472 | 0.199 |
| corridor5 | **0.227** | × | 0.499 | **0.262** | × | 0.384 |
| slides1 | **0.261** | × | 0.486 | **0.325** | 0.901 | 0.338 |
| slides2 | 1.214 | × | **1.052** | **0.496** | × | 0.949 |
| slides3 | **0.446** | × | 0.628 | **1.149** | 2.192 | 1.769 |
| magistrale1 | **2.143** | × | 3.582 | 2.441 | × | **1.720** |
| magistrale2 | **1.313** | 4.080 | 2.409 | **1.230** | 2.962 | 1.274 |
| magistrale3 | 2.577 | × | **0.668** | 2.354 | **0.812** | 1.964 |
| magistrale4 | **1.261** | 2.429 | 4.374 | 2.311 | 1.773 | **1.352** |
| magistrale5 | 1.520 | × | **0.225** | 1.392 | × | **0.955** |
| magistrale6 | **2.139** | × | 2.763 | **2.084** | 3.622 | 3.737 |
| Mean (common seq.) | 0.377 | 0.851 | 1.019 | 0.718 | 1.022 | 0.819 |
| Completed | 20/20 | 11/20 | 20/20 | 20/20 | 15/20 | 20/20 |

### C. 360° Camera

The comparison systems in Table I do not support stitched 360° ERP input, so we evaluate OpenVIO alone on PanoAir [1]. It contains 17 drone flights totaling 15.5 km under cloudy, sunny and night conditions, with RTK-GNSS ground truth. Images from a two-lens camera have an ERP resolution of 1280 × 640 at 30 Hz. We use one run per flight with the released configuration and exclude the first 12.5 s from evaluation. OpenVIO produces a successful result on all 17 flights, with a mean ATE of 1.14 m (Table V).

Table V also reports the final online calibration estimates. No reference values are available for these calibration parameters. The stitching parameter $k$ lies between 0.0055 and 0.0068 on 15 flights, with flights 7 and 13 outside this range. The estimated back-lens delay has magnitude below 0.5 ms on every flight except flight 10. For that flight, the estimate is 35 ms, approximately one frame interval at 30 Hz.

## V. Conclusion

OpenVIO is an MIT-licensed visual-inertial odometry system for pinhole, fisheye and stitched 360° ERP images. A shared camera interface supports these models within one filter, with online correction of stitching error and inter-lens exposure delay for two-lens 360° cameras. OpenVIO meets the benchmark success criterion on all evaluated EuRoC and TUM-VI sequences in monocular and stereo modes, and on all 17 PanoAir flights. Loop closure and map reuse are not included; exported keyframes and feature observations provide an interface for such extensions.

TABLE V

PanoAir: ATE RMSE [m] of OpenVIO and the online calibration at the end of each flight

| | Cloudy | | | | | Sunny | | | | Night | | | | | | | | |
|---|---|---|---|---|---|---|---|---|---|---|---|---|---|---|---|---|---|---|
| Flight | 1 | 2 | 3 | 4 | 5 | 6 | 7 | 8 | 9 | 10 | 11 | 12 | 13 | 14 | 15 | 16 | 17 | Mean |
| Length [m] | 248 | 802 | 1143 | 1206 | 1591 | 792 | 500 | 839 | 1700 | 364 | 472 | 407 | 464 | 831 | 903 | 1355 | 1883 | |
| ATE | 0.29 | 0.61 | 1.08 | 1.90 | 2.99 | 0.39 | 0.36 | 0.58 | 1.71 | 1.39 | 0.81 | 0.74 | 0.53 | 1.20 | 1.61 | 1.47 | 1.74 | 1.14 |
| $k \times 10^3$ | 6.31 | 6.67 | 6.34 | 6.17 | 6.46 | 6.43 | 3.62 | 6.39 | 6.73 | 6.61 | 5.88 | 5.59 | 8.76 | 5.78 | 6.05 | 5.50 | 6.58 | |
| $t_d$ [ms] | −1.8 | −4.1 | −4.2 | −3.8 | −4.7 | −6.3 | −3.3 | −5.1 | −6.8 | −39.0 | −4.3 | −3.5 | −3.6 | −5.2 | −4.6 | −5.4 | −6.9 | |
| $\delta t_b$ [ms] | 0.31 | 0.27 | 0.15 | 0.15 | 0.41 | 0.16 | 0.06 | 0.22 | 0.15 | 35.00 | −0.17 | −0.18 | −0.15 | −0.13 | −0.16 | 0.02 | −0.20 | |

One run per flight with the released configuration; the first 12.5 s of each flight are excluded from evaluation. Calibration: final online estimates of the stitching parameter $k$ (prior $(6.3 \pm 1.5) \times 10^{-3}$), the camera time offset $t_d$ and the back-lens delay $\delta t_b$. The $k$ prior uses earlier online estimates from this dataset.